\documentclass[journal]{IEEEtran}
\ifCLASSINFOpdf
   \usepackage[pdftex]{graphicx}
\else
\fi
\ifCLASSOPTIONcompsoc
  \usepackage[caption=false,font=normalsize,labelfont=sf,textfont=sf]{subfig}
\else
  \usepackage[caption=false,font=footnotesize]{subfig}
\fi

\usepackage{amsmath}
\usepackage{amssymb}

\newcommand{\transpose}{{\mbox{\scriptsize T}}}

\newcommand{\N}{\mathcal{N}}
\newcommand{\E}{\mathbb{E}}
\newcommand{\bs}{\boldsymbol}

\newcommand{\by}{\textbf{y}}
\newcommand{\bY}{\textbf{Y}}

\newcommand{\bV}{\textbf{V}}

\newcommand{\bff}{\textbf{f}}

\usepackage{multirow}

\begin{document}
%
\title{Multi-output Gaussian processes\\ for crowdsourced traffic data imputation}
%
%
%

\author{Filipe~Rodrigues,
		Kristian~Henrickson,
        and~Francisco~C.~Pereira,~\IEEEmembership{Member,~IEEE}
\thanks{F.~Rodrigues is with the Technical University of Denmark (DTU), Bygning 115, 2800 Kgs. Lyngby, Denmark. E-mail: rodr@dtu.dk}
\thanks{K.~Henrickson is with the University of Washington, 3760 E. Stevens Way NE, 98195 Seattle, WA, USA.}
\thanks{F.~C.~Pereira is with the Technical University of Denmark (DTU), Bygning 115, 2800 Kgs. Lyngby, Denmark, and the Massachusetts Institute of Technology (MIT), 77 Massachusetts Avenue, 02139 Cambridge, MA, USA.}
\thanks{DOI: https://doi.org/10.1109/TITS.2018.2817879}\thanks{URL: https://ieeexplore.ieee.org/document/8352762}}

%
%

\markboth{IEEE Transactions on Intelligent Transportation Systems, 2018}%
{Rodrigues \MakeLowercase{\textit{et al.}}: Multi-Output Gaussian Processes for Crowdsourced Traffic Data Imputation}
%



\maketitle

\begin{abstract}
Traffic speed data imputation is a fundamental challenge for data-driven transport analysis. In recent years, with the ubiquity of GPS-enabled devices and the widespread use of crowdsourcing alternatives for the collection of traffic data, transportation professionals increasingly look to such user-generated data for many analysis, planning, and decision support applications. However, due to the mechanics of the data collection process, crowdsourced traffic data such as probe-vehicle data is highly prone to missing observations, making accurate imputation crucial for the success of any application that makes use of that type of data. 

In this article, we propose the use of multi-output Gaussian processes (GPs) to model the complex spatial and temporal patterns in crowdsourced traffic data. While the Bayesian nonparametric formalism of GPs allows us to model observation uncertainty, the multi-output extension based on convolution processes effectively enables us to capture complex spatial dependencies between nearby road segments. Using 6 months of crowdsourced traffic speed data or ``probe vehicle data" for several locations in Copenhagen, the proposed approach is empirically shown to significantly outperform popular state-of-the-art imputation methods.
\end{abstract}

\begin{IEEEkeywords}
Traffic data, imputation, missing data, Gaussian processes, multiple outputs, Bayesian inference, crowdsourced data.
\end{IEEEkeywords}

%
\IEEEpeerreviewmaketitle

\section{Introduction}
%
%
%
%
\IEEEPARstart{M}{issing} data imputation is an important consideration in a wide variety of science and engineering fields, and as such has been the focus of a great number of research publications in the past decades. Traffic data imputation in particular is an active area of research, which until recent years has largely focused on data collected through fixed mechanical sensors such as loop detectors. However, due in part to the pervasiveness of GPS-capable smartphones, probe vehicle data or data collected from mobile GPS devices has been growing in popularity and gradually augmenting or even replacing conventional data sources in many transport applications (e.g. \cite{gong2017applying,mauch2015validating}). As a result, there is clear need for imputation methods that can address the missing data challenges unique to this data collection paradigm that is based on crowdsourcing. 

The prevailing view in missing data science categorizes missing data patterns as Missing Completely At Random (MCAR), Missing At Random (MAR) or Not Missing At Random (NMAR) \cite{little2014statistical}. MAR describes a scenario where the probability of a particular observation being missing is independent of all unobserved information conditioned on the observed values. MCAR is a special case of MAR, and indicates that the probability of an observation being missing is completely independent of both observed and unobserved information. The most challenging scenario is NMAR, which describes the case where the probability of missingness is related to some unobserved information (including the missing values themselves). MAR and MCAR can be described as arising from an ignorable missing data mechanism, in that the missing data mechanism can be ignored in developing an imputation model. Much of the previous work on missing data imputation has targeted ignorable missing data and it is clear that, especially when the true missing data mechanism is not well understood, there is a strong motivation to treat data as MAR even when this assumption is not well supported. In such cases, an imputation procedure should incorporate as much information that may help to describe the difference between the missing and observed data as possible in order to make the MAR assumption tenable and minimize the potential bias associated with the missingness pattern \cite{collins2001comparison,little2014statistical}. This is particularly relevant to probe vehicle data where, unlike data from fixed mechanical sensors, previous research suggests that the probability of an observation being missing is often related to traffic state i.e. volume and/or speed \cite{henrickson2016}. 

A great variety of regression approaches have been applied to the task of traffic data imputation in the past, though most were developed to address missing observations in loop detectors and other fixed mechanical sensors rather than probe vehicle data. Common modeling approaches include linear regression \cite{chen2003detecting}, support vector machines \cite{zhang2009data}, K-nearest neighbors and other non-parametric methods \cite{chang2012missing,li2010incident,liu2008imputation}, and others \cite{tang2015hybrid,zhong2004genetically}. Expectation Maximization methods have also been used to some extent in traffic volume and speed imputation, with mixed results \cite{chang2011comparison,dempster1977maximum,tak2016data}. The primary challenge associated with conventional regression methods is in identifying and specifying an appropriate model that can capture the complex dynamics present in traffic data, and incorporating relevant spatial and temporal features in order to achieve accurate, unbiased imputations.  

Matrix and tensor completion methods are growing increasingly popular for traffic data imputation and prediction \cite{mardani2015subspace,ran2016using,tan2013traffic,chiou2014functional}. Such methods have been shown to provide a reasonably accurate and computationally efficient framework for imputing traffic datasets, though the focus has been primarily on minimizing large-scale reconstruction error rather than quantifying uncertainty. Other dimension reduction techniques, such as probabilistic principle component analysis (PPCA), have also been applied to traffic data imputation \cite{li2013efficient,qu2009ppca}. Some extensions have been proposed which seek to overcome the limitations of conventional PPCA (e.g. \cite{zhao2014improving}), specifically assumptions regarding the linear relationship between the observed and latent variables, the requirement that latent variables follow a single Gaussian distribution, and challenges in incorporating temporal dependencies \cite{li2013comparison,zhang2011data}. 

Several publications have developed complex neural network or deep learning models for traffic state prediction and, to a lesser extent, missing data imputation \cite{duan2016efficient,elhenawyspatiotemporal,ma2017learning,yu2017spatiotemporal}. Such methods have been shown to provide accurate predictions, especially for complex, high dimensional traffic data, though no previous work has applied this class of models to the task of probe vehicle-based link-level speed or travel time data completion. With a focus on predictive power rather than model interpretability or quantifying uncertainty \cite{beaulieu2016missing,gal2016uncertainty,ma2017learning}, much of the current generation of deep learning methods may not be appropriate data imputation tools for in-depth statistical inquiry.

Multiple imputation (MI), has been applied to traffic data in some previous work \cite{henrickson2014flexible,ni2005multiple,henrickson2015flexible}. By drawing from some predictive distribution, $m > 1$ replacement values are generated for each missing observation resulting in m complete datasets which capture the uncertainty attributable to missing data. The benefit of MI is in the ability to produce both unbiased point estimates to replace missing values and robust uncertainty measures. MI represents a very general framework for treating missing data and quantifying uncertainty, and has been applied to traffic data with a range of modeling approaches including semi and non-parametric regression models \cite{henrickson2014flexible,henrickson2015flexible}, and Bayesian methods \cite{kwon2004tmc,ni2005multiple}. However, although not due to any specific limitations of the MI framework, most previous work on traffic data imputation using MI has not considered spatial and temporal dependencies in tandem for predicting missing values. 

This article proposes the use of multi-output Gaussian processes \cite{boyle2004dependent,alvarez2011computationally} in order to model the complex spatial and temporal patterns in traffic data. Gaussian processes (GPs) are flexible non-parametric Bayesian models that are widely used for modeling complex time-series. Indeed, single-output GPs have been successfully applied to model and predict with state-of-the-art results various traffic related phenomena such as traffic congestion \cite{liu2013adaptive}, travel times \cite{ide2009travel,rodrigues2016bayesian}, pedestrian and public transport flows \cite{neumann2009stacked}, traffic volumes \cite{xie2010gaussian}, etc. As pointed out in \cite{li2015trend}, one of the key benefits of multi-output Gaussian processes lies in its potential to better capture the varying trend and stochastic features of traffic flow. Additionally, the fully Bayesian non-parametric formulation of GPs makes them particularly well suited for modeling uncertainty and noise in crowdsourced traffic data. Moreover, the multi-output formulation of GPs described in this article allows to accurately model complex dependencies between the traffic conditions at nearby road segments. 

The approach described in this article therefore addresses several shortcomings of existing work on probe vehicle data imputation. First, by developing a fully probabilistic framework for imputation, we are able to characterize the full conditional distribution of the missing values and describe the uncertainty associated with the imputed values. With some exceptions, most previous work in statistical and machine learning methods for traffic data imputation has focused on minimizing reconstruction error, and largely ignored the need to accurately quantify uncertainty in the imputed values. Second, our approach is able to capture the complex spatial and temporal dependencies present in traffic data. As noted previously, this is needed to support the MAR assumption and particularly relevant to probe vehicle data. Much of the previous work on traffic data imputation was developed for fixed mechanical sensing data which, as discussed previously, is most often subject to less complex missing data patterns. Finally, by applying Gaussian Process regression in a multi-output framework, the proposed methodology is able to better preserve the relationship between speed records obtained at adjacent road segments under moderate to high rates of missingness. As our experimental results show, the proposed approach is highly effective in traffic speed data imputation and it significantly outperforms popular state-of-the-art alternatives. 

The rest of this article is organized as follows. First, Section~\ref{sec:gp} introduces GPs and discusses how to use them for modeling time-series data. The proposed approach based on multi-output Gaussian processes is presented in Section~\ref{sec:multigp}. A thorough experimental evaluation of the proposed methodology in comparison with other state-of-the-art approaches is presented
in Section~\ref{sec:experiments}. Finally, we conclude in Section~\ref{sec:conclusion}.

\section{Gaussian processes for traffic speeds}
\label{sec:gp}

In this section, we describe how to model a time-series $\by_r = \{y_{r,1},...,y_{r,T}\}$ of $T$ observations of speeds in a single road segment $r$, independently of the others, using Gaussian processes. This will be the basis for the extension to multiple outputs described in Section~\ref{sec:multigp}.

Let the observed speed in a road segment $r$ at time $t$ be defined as
\begin{align}
y_{r,t} = f_r(t) + \epsilon_r,
\label{eq:homeskedastic_formulation}
\end{align}
where $f_r$ is an unknown function of time and $\epsilon_r$ is an additive white noise process, such that $\epsilon_r \sim \N(\epsilon_r|0,\sigma_r^2)$. Gaussian process (GP) approaches to time-series modeling assume $f_r$ to be a non-linear non-parametric function and proceed by placing a GP prior over $f_r$. 

A Gaussian process is a stochastic process fully specified by a mean function $m_r(t) = \E[f_r(t)]$ and a positive definite covariance function $k_r(t,t') = \mbox{cov}[f_r(t),f_r(t')]$. By making use of the mean and covariance functions, GPs specify a way to determine the mean of any arbitrary point in time $t$ and how that point covaries with other points in time. Therefore, we can think of a Gaussian process as a probability distribution over functions $f_r$. By placing a GP prior over $f_r$ with certain mean and covariance functions, we are then specifying our prior knowledge about the properties of the observed time-series of speeds. 

\begin{figure*}[ht]
\centering
\subfloat[squared exponential (SE)]{\includegraphics[width=0.25\linewidth]{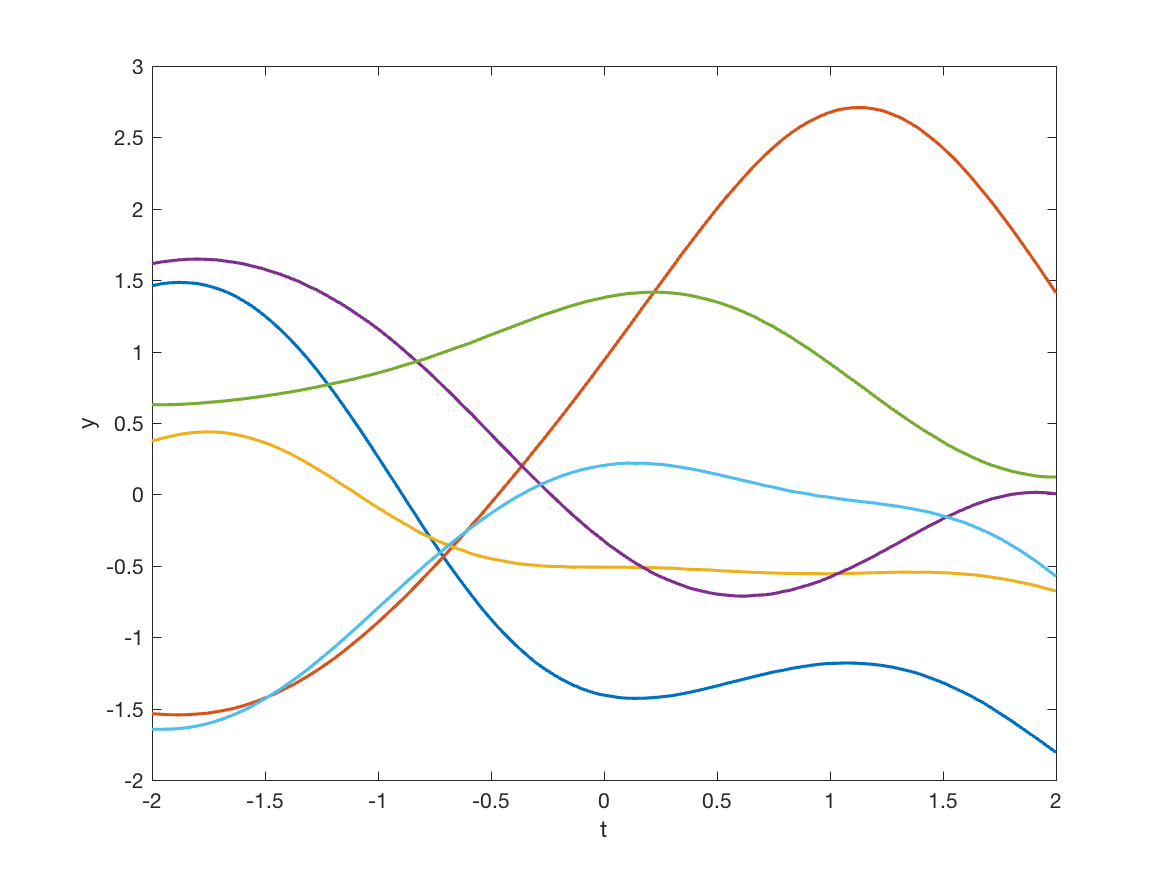}\label{fig:gp_sample_SE}}\hspace{-0.25cm}
\subfloat[periodic (PER)]{\includegraphics[width=0.25\linewidth]{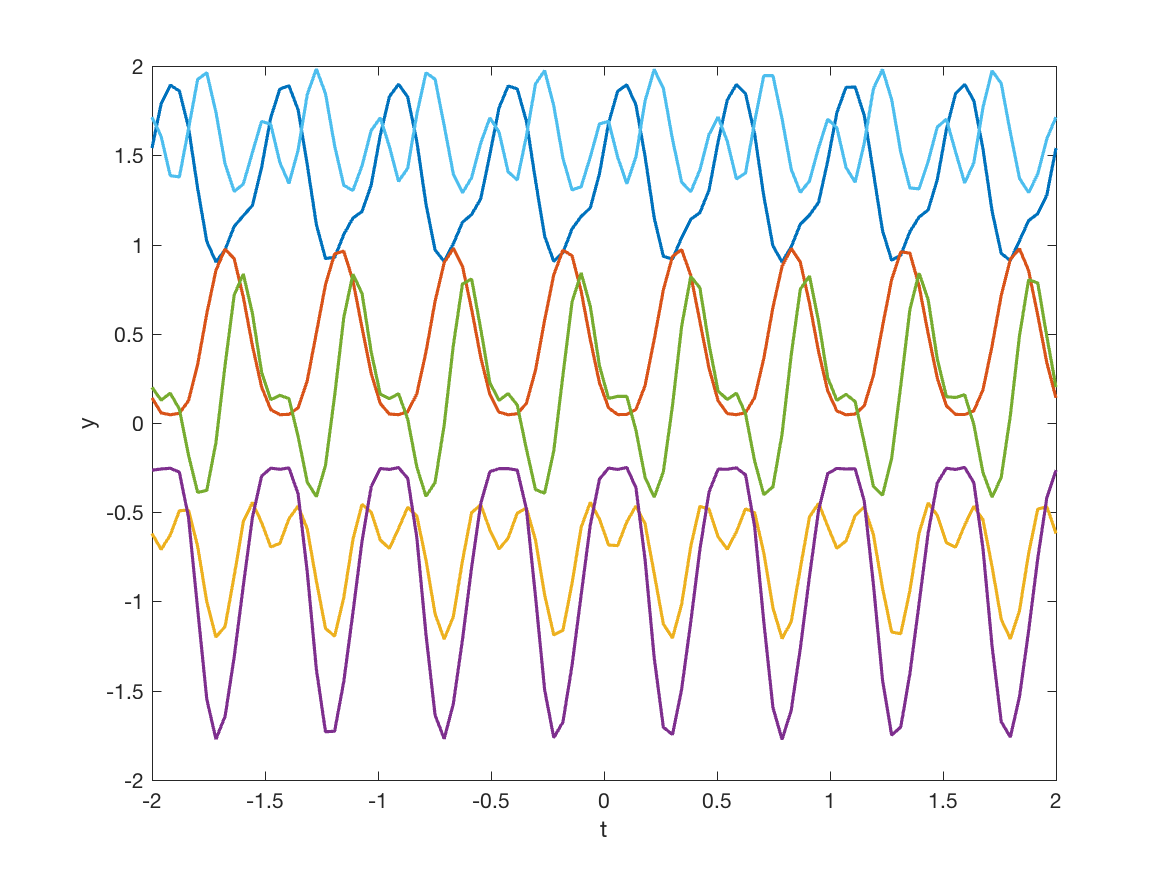}\label{fig:gp_sample_PER}}\hspace{-0.25cm}
\subfloat[white noise (WN)]{\includegraphics[width=0.25\linewidth]{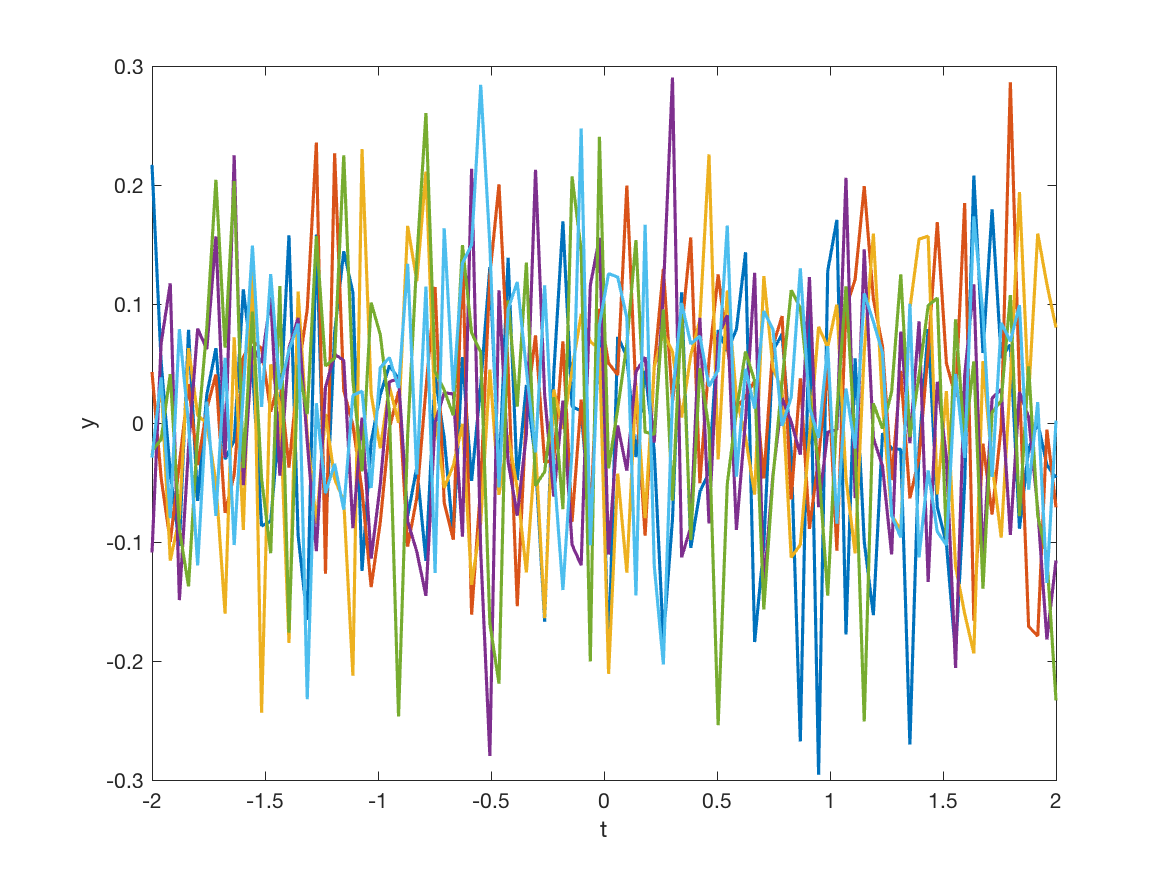}\label{fig:gp_sample_WN}}\hspace{-0.25cm}
\subfloat[SE+WN]{\includegraphics[width=0.25\linewidth]{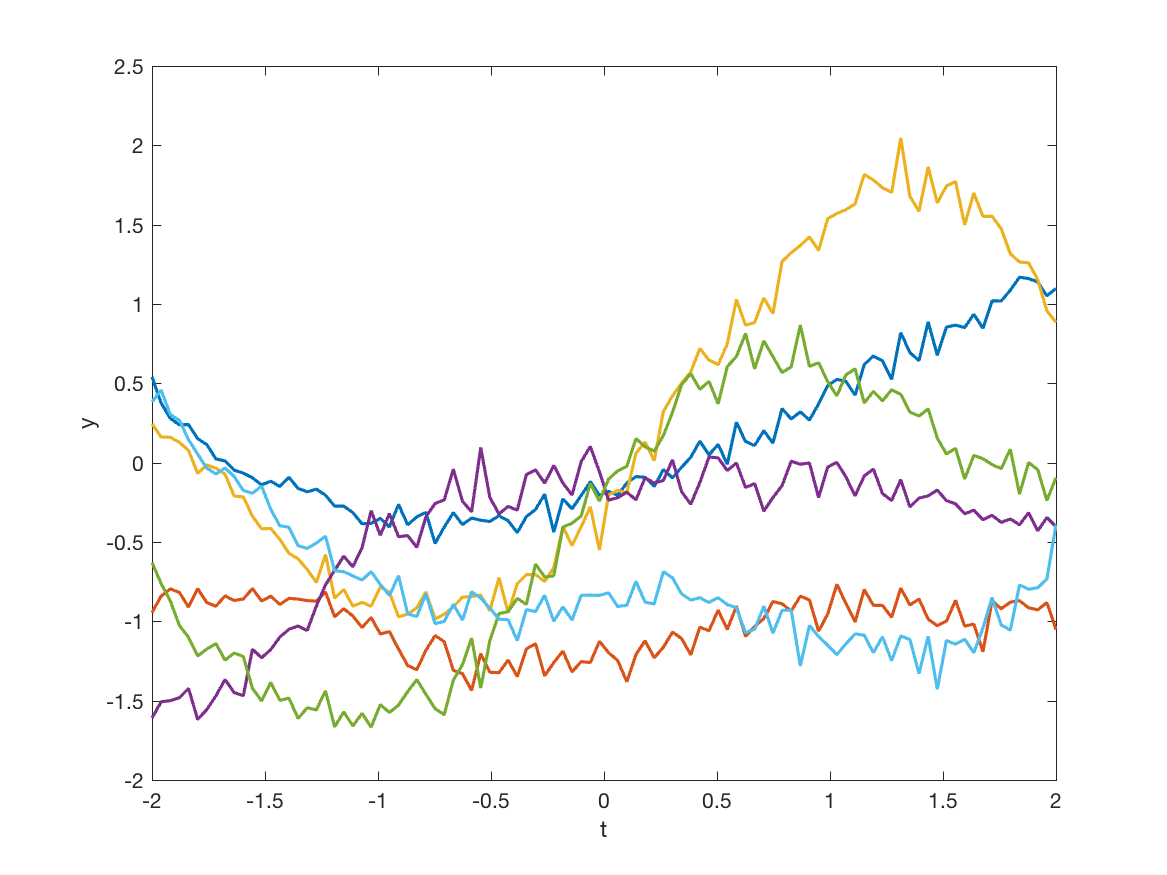}\label{fig:gp_sample_SE+WN}} 
\caption{Samples from Gaussian process priors with different covariance functions.}
\end{figure*}

Given proper normalization of time-series data, we can assume the mean of the process to be zero, \mbox{i.e.} $m_r(t) = \E[f_r(t)] = 0$. As for the covariance function, it specifies basic aspects of the process, such as stationarity, isotropy, smoothness and periodicity. A fundamental property of temporal data, such as traffic speeds, is that speed observations at nearby time indexes tend to be similar. We can elegantly incorporate this property in Gaussian processes by using a squared exponential (SE) covariance functions, which is defined as
\begin{align}
k_{\mbox{\tiny SE}}(t,t') = h^2 \, \exp \bigg( - \frac{(t-t')^2}{2\ell^2} \bigg), 
\label{eq:sq_exp}
\end{align}
where the parameter $\ell$ defines the characteristic length-scale and $h$ specifies an output-scale amplitude. Notice how the exponent goes to unity as $t$ becomes closer to $t'$. Hence, nearby points are more likely to covary. As a result, a GP prior with a squared exponential covariance function prefers smooth functions. Figure~\ref{fig:gp_sample_SE} shows 5 samples from this covariance function. It is important to note that sums and products of valid covariance functions also produce valid covariance functions. Therefore, we can incorporate additional properties of the time-series in the GP simply by adding covariance functions. 

Another important property of traffic speeds is periodicity. This property can be incorporated into a GP by making use of a periodic (PER) covariance function, given by
\begin{align}
k_{\mbox{\tiny PER}}(t,t') = h^2 \, \exp \bigg( - \frac{1}{2\ell^2} \sin^2 \bigg(\pi \frac{(t-t')^2}{p} \bigg) \bigg), \nonumber
\end{align}
where $h$ controls the amplitude and $p$ is the period. Lastly, we can account for uncertainty in the observed speeds by including a white noise (WN) covariance function with variance $\sigma^2$, which is defined as
\begin{align}
k_{\mbox{\tiny WN}}(t,t') = \sigma^2 \, \delta(t,t'), \nonumber
\end{align}
where $\delta(t,t')$ is the Kronecker delta function which takes the value 1 when $t = t'$ and 0 otherwise. Figures~\ref{fig:gp_sample_PER} to \ref{fig:gp_sample_SE+WN} show samples from these covariance functions.

One major advantage of the GP formulation described above is that exact Bayesian inference is tractable. Let $\bff_r = \{f_r(t)\}_{t=1}^T$ denote the function $f_r(t)$ evaluated for different time indexes $t$. From (\ref{eq:homeskedastic_formulation}), and assuming \mbox{i.i.d.} observations, we can write the likelihood as $\by_r|\bff_r \sim \N(\by_r|\bff_r,\sigma_r^2\textbf{I})$, where $\textbf{I}$ refers to the identity matrix. Having specified a GP prior for the function values, $p(\bff_r) = \mathcal{GP}(m_r(t)=0, k_r(t,t'))$, the posterior distribution of the speed $y_{r,*}$ in road segment $r$ for an unobserved time $t_*$, which we wish to impute, can be computed in closed form by making use of Bayes' theorem, yielding \cite{rasmussen2006gaussian}
\begin{align}
p(y_{r,*}|\by_r) &= \N(y_{r,*}|\textbf{k}_{r,*}^\transpose \bV_r^{-1} \by, \, k_{r,**} + \sigma_r^2 - \textbf{k}_{r,*}^\transpose \bV_r^{-1} \textbf{k}_{r,*}),
\label{eq:pred_dist_exact}
\end{align}
where $\bV_{r} \triangleq \sigma_r^2\textbf{I} + \textbf{K}_{r}$, with the $T \times T$ matrix $\textbf{K}_r$ consisting of the covariance function $k_r(t,t')$ evaluated between every pair of time indexes, $\textbf{k}_{r,*}$ denotes the covariance function evaluated between the imputation time $t_*$ and all the other observed time indexes, and $k_{r,**}$ denotes the covariance function evaluated between the imputation time $t_*$ against itself, i.e. $k_{r,**} = k_r(t_*,t_*)$.

One key advantage of the Bayesian formalism of GPs is the fully probabilistic interpretation of the predictions and their ability to handle uncertainty, which can be verified by plotting the predictive distributions in (\ref{eq:pred_dist_exact}) and noticing that the uncertainty is lower close to the observations and becomes higher as we go towards regions with no observations. Therefore, GPs effectively allow us to capture the uncertainty in our inferences about the missing observations. 

So far we have been assuming the hyper-parameters of the covariance function $k_r(t,t')$ to be fixed. However, these can be optimized by maximizing the logarithm of the marginal likelihood of the observations given by
\begin{align}
p(\by_r) &= \int p(\by_r|\bff_r) \, p(\bff_r) \, d\bff_r  = \N(\by_r|\textbf{0},\bV_r).
\label{eq:marg_lik_exact}
\end{align}

\section{Spatio-temporal modeling of speeds with multi-output GPs}
\label{sec:multigp}

In the previous section, we described a GP approach for modeling traffic speeds that exploits temporal correlations and other properties of the time-series in order to infer missing observations. However, besides the temporal correlations captured by the temporal process $f_r(t)$, traffic speeds are known to also exhibit strong spatial correlations between adjacent (or nearby) road segments. We therefore would like to leverage these correlations and dependencies between road segments in order to more accurately infer the unobserved speeds. A key difficulty in formulating multi-output GPs for jointly modeling the traffic speeds at different road segments lies in specifying an appropriate covariance function that captures the dependencies between all time indexes and across all road segments, while also leading to positive semi-definite covariance matrices. 

An elegant way for accounting for non-trivial correlations between the outputs of different Gaussian processes is based on the convolution processes (CP) formalism \cite{boyle2004dependent,alvarez2011computationally}. Let $\{s_r(t)\}_{r=1}^R$ be a set of $R$ correlated spatial speed functions for different road segments $r$, which we wish to model. Under the CP formalist, we can express $s_r(t)$ as a convolution of an input process $u(z)$, such as a white noise process, with a smoothing kernel function $k_r(t-z)$, 
\begin{align}
s_r(t) &= \int k_r(t-z) \, u(z) \, dz.
\end{align}
As noted in \cite{ver1998constructing}, if the same input process $u(z)$ is convolved with different smoothing kernels to produce different outputs, then correlation between outputs can be expressed. More generally, we shall consider a set of $Q$ latent functions $\{u_q(z)\}_{q=1}^Q$. The correlated observed speeds $y_{r,t}$ in a road segment $r$ at time $t$ can then be defined as
\begin{align}
y_{r,t} &= s_r(t) + f_r(t) = \sum_{q=1}^Q \int k_{rq}(t-z) \, u_q(z) \, dz + f_r(t),\nonumber
\end{align}
where $f_r(t)$ is an independent temporal process as described in the previous section. Therefore, while the temporal process $f_r(t)$ captures the temporal structure in the speed time-series of each road segment $t$, the spatial process $s_r(t)$ is responsible for capturing correlations with nearby road segments. 

Since the convolution is a linear operator on a function, the outputs of the convolutions can be expressed as a jointly distributed GP. The covariance between the speeds at two road segments, $r$ and $h$, is then given by
\begin{align}
\mbox{cov}[y_{r,t},y_{h,t'}] &= \mbox{cov}[s_r(t),s_h(t')] + \delta(t,t') \, \mbox{cov}[f_r(t),f_h(t')],
\label{eq:cov_multi}
\end{align}
where
\begin{align}
\mbox{cov}[s_r(t),s_h(t')] &= \sum_{q=1}^Q \sum_{p=1}^Q \int k_{rq}(t-z) \int k_{hp}(t'-z')  \nonumber\\ 
&\times \mbox{cov}[u_q(z),u_p(z')] \, dz' \, dz.
\label{eq:cov_multi_func1}
\end{align}
Further assuming that the latent functions $u_r(z)$ are independent GPs, such that $\mbox{cov}[u_q(z),u_p(z')] = \delta(z,z') \, k_{u_r u_p}(z,z')$, where $k_{u_r u_p}(z,z')$ is the covariance function for $u_r(z)$, we can simplify (\ref{eq:cov_multi_func1}) as
\begin{align}
&\mbox{cov}[s_r(t),s_h(t')] \nonumber\\ &= \sum_{q=1}^Q \int k_{rq}(t-z) \int k_{hq}(t'-z') \, k_{u_r u_p}(z,z') \, dz' \, dz.
\label{eq:cov_ss2}
\end{align}
As long as the integral in (\ref{eq:cov_ss2}) has a closed form solution (which is the case for Gaussian kernels such as the one in (\ref{eq:sq_exp}); see \cite{boyle2004dependent}), the CP formalism allows us to build positive semi-definite covariance matrices for jointly modeling the traffic speeds $\by_r$ at different correlated road segments $r$. 

Let $\by = [\by_1^\transpose,...,\by_R^\transpose]^\transpose$ be a set of $R$ related time-series of traffic speeds, typically for spatially close road segments. As with the independent case from Section~\ref{sec:gp}, and without loss of generality, we shall assume the mean function of the multi-output GP to be zero. The marginal likelihood for a full GP over $\bY$ is then given by
\begin{align}
p(\bY) &= \N(\bY|\textbf{0},\textbf{K}+\bs\Sigma),
\end{align}
where $\textbf{K} \in \mathbb{R}^{RT \times RT}$ is a covariance matrix obtained by evaluating the covariance function in (\ref{eq:cov_multi}) between all the observations from all the road segments, and $\bs\Sigma$ is a block diagonal matrix with diagonal blocks $\bs\Sigma_t = \mbox{diag}(\sigma_1^2,\dots,\sigma_R^2)$. 

The joint predictive distribution for the traffic speeds in all road segments at time $t_*$ is then given by  \cite{rasmussen2006gaussian}
\begin{align}
p(\by_*|\by) &= \N(\by_*|\textbf{K}_{*}^\transpose \bV^{-1} \by, \, \textbf{k}_{**} + \bs\Sigma - \textbf{K}_{*}^\transpose \bV^{-1} \textbf{K}_{*}),
\label{eq:posterior_multi}
\end{align}
where $\bV \triangleq \textbf{K} + \bs\Sigma$, the matrix $\textbf{K}_{*}$ is obtained by evaluating the covariance function in (\ref{eq:cov_multi}) between time $t_*$ in all road segments and all the other observations from all segments, and analogously for the vector $\textbf{k}_{**}$. 

\section{Experiments}
\label{sec:experiments}

\begin{table*}[t]
\centering
\caption{Descriptive statistics of study areas.}
\label{table:descriptive_stats}
\begin{tabular}{lllllll}
\hline
Area & Num. Lanes & Road type & Traffic lights & Speed limit & Average speed & Speed std. dev.\\
\hline
1 & 2 (+1 Bus) & Arterial with BRT & 100m before & 50 km/h & 23.3 km/h & 16.9 km/h \\
2 & 2 (+1 Bus) & Arterial with BRT & 100m after & 50 km/h & 30.3 km/h  & 18.3 km/h \\
3 & 2 & Local road & 150m after & 50 km/h & 21.8 km/h  & 16.3 km/h \\
4 & 2 (+1 Bus) & Intersection & Just before & 50 km/h & 29.0 km/h & 15.5 km/h \\
\hline
\end{tabular}
\end{table*}

\begin{figure}[t]
	\centering
	\includegraphics[width=1\linewidth]{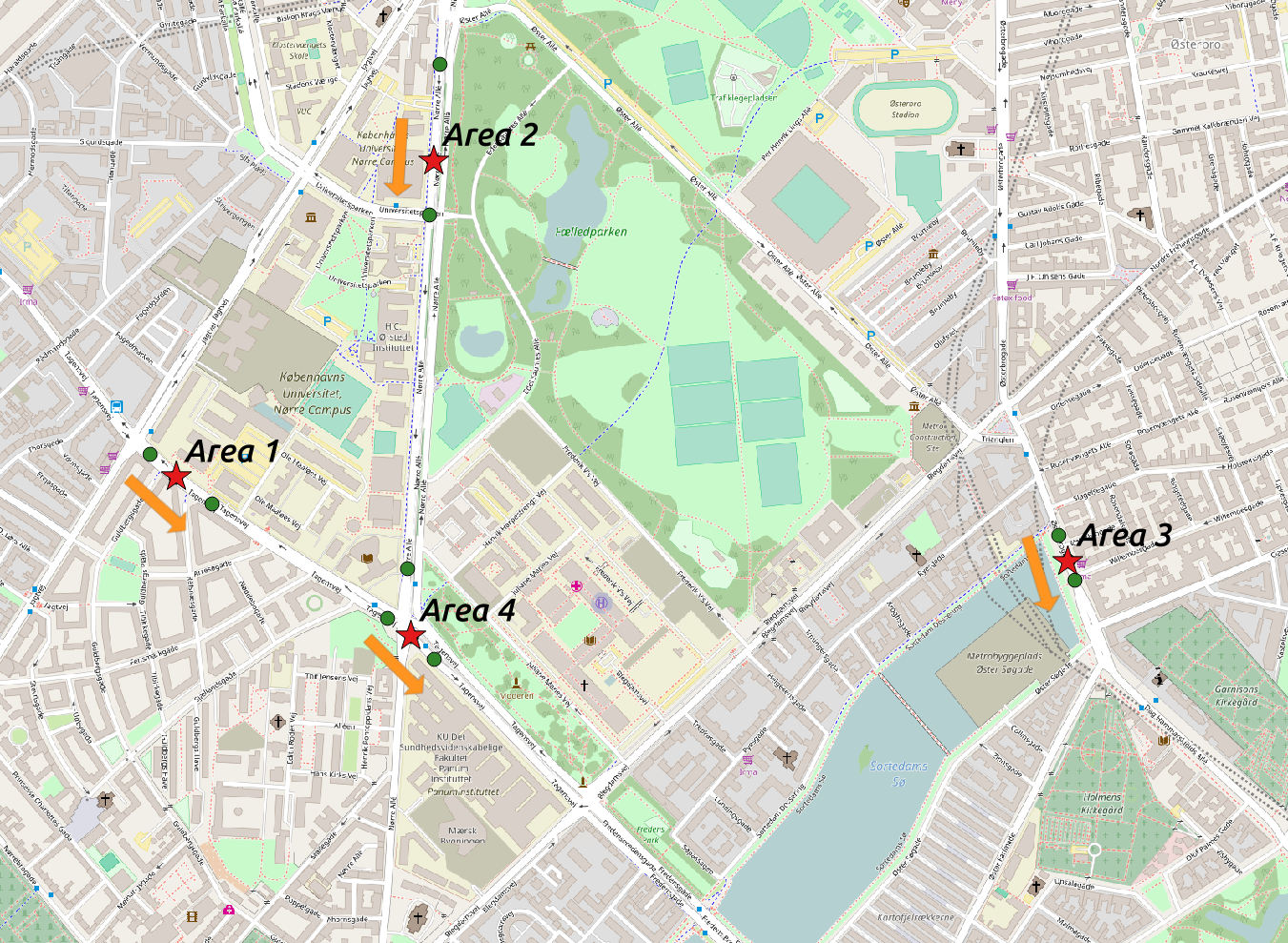}
	\caption{Map of the 4 study areas. Red stars denote the target road segments whose missing speeds we want to impute, and green circles represent nearby road segments. Arrows indicate the direction of traffic at the road segments.}
	\label{fig:map}
\end{figure}

In this section, we empirically evaluate the performance of the proposed multi-output Gaussian process for traffic speed imputation\footnote{Source code provided as supplementary material.} in different settings and in comparison with popular imputation methods from the state of the art. We begin by introducing the dataset used in the experiments. Then, we analyze the importance of exploring the dependencies between traffic speeds at nearby road segments, and we evaluate the performance of the different imputation approaches for different percentages of missing data. Lastly, we analyse the effect of different missing patterns in the results. 

\subsection{Dataset}

The data used to evaluate the proposed approach consists of 6 months (January 2015 to June 2015) of crowdsourced traffic speeds provided by Google for 13 road segments distributed across 4 study areas in Copenhagen. Figure~\ref{fig:map} shows a map of the study areas, each consisting of target road segments (red star) whose missing speeds we wish to impute, and adjacent road segments (green circles) whose relation we want to explore for improving imputation accuracy at the target road segments. Table~\ref{table:descriptive_stats} shows descriptive statistics of the 4 study areas considered. Kindly notice how they are representative of different types of scenarios, such as arterials with bus rapid transit (BRT), major intersections and smaller local roads. 

The crowdsourced traffic data provided by Google is derived from “Location History” data shared by Google Maps users which have that option activated. Location History is the same data used in Google Now to notify users of disruptions to their commute due to traffic and recommend the best time to visit attractions or restaurants in Google Maps. The individual GPS data from the mobile users is then aggregated per road segment in 5 minute bins. The road segments are predefined by Google and uniquely identified using unique place IDs, whose details can be obtained through the use of the Google's Places API.

\subsection{Importance of multi-output methods}

With the purpose of studying the effect of jointly modeling the traffic speeds at nearby road segments and the importance of exploiting the correlations between those segments, we analyzed the performance of the proposed multi-output GP in different settings. Namely, we consider jointly modeling traffic data from the previous and subsequent road segments in order to improve imputation at the target road segments of each study area. In order to simulate missing data, we randomly select a fraction of the observations (MCAR assumption) and remove them from the dataset. The goal is then to use the remaining data to infer the missing observations. 

The proposed multi-output GP approach (referred to as ``Multi-GP") is compared with the following baselines: 
\begin{itemize}
\item A naive method that simply predicts the previously observed speed (referred to as ``Naive");
\item An autoregressive integrated moving average model (ARIMA);
\item A linear regression model that uses the 5 previous and 5 next observed speeds, as well as the speeds at other road segments, to infer the traffic speed in the target area at time $t$ (referred to as ``Lin. Reg.");
\item A k-nearest neighbours (kNN) approach that uses the inverse Euclidian distance to weight the $k$ closest neighbours (as used in \cite{tak2016data});
\item A bootstrap-based Expectation-Maximization (B-EM) approach from \cite{honaker2011amelia}, which assumes a multivariate Gaussian distribution of data and uses a bootstrap sampling method and Expectation-Maximization algorithm \cite{dempster1977maximum} in order to perform imputation;\footnote{It is important to note that this approach is radically different from the our proposed approach based on multi-output GPs, which leverages the use of convolution processes (CP) and the properties of the covariance function in order to capture complex spatio-temporal patterns in the data and impute missing observations through full posterior (Bayesian) inference.}
\item A matrix factorization approach based on Probabilistic PCA (PPCA) that explores spatio-temporal correlations in the data as proposed in \cite{qu2009ppca,li2013efficient}. In order to explore the spatio-temporal correlations, the data from different days for different road segments is put together in a single matrix as described in \cite{li2013efficient}, \mbox{Eq.} 24;
\item A deep learning approach based on recurrent neural networks (RNNs), namely a Bidirectional Long Short-term Memory (Bidirectional LSTM) \cite{graves2013hybrid} with an additional LSTM layer stacked on top and, finally, a fully connected dense output layer with linear activations; 
\item An independent GP using only the temporal structure at the target road segment, as described in Section~\ref{sec:gp} (referred to as ``Indep. GP");
\item A vector autoregressive moving average model (VARMA).
\end{itemize}
The orders of the ARIMA and VARMA models were automatically determined using the Akaike information criterion (AIC).\footnote{We experimented with other criteria such as BIC, as well as different manually specified orders. For a wide majority of reasonable choices, the differences in imputation accuracy were not significant.} Also, please note that both these methods effectively use data from both the past and the future to perform imputation of missing values. The number of latent functions used in the multi-output GPs is equal to the number of outputs. 

In order to evaluate the performance of the different methods from different perspectives, we report the following error metrics: mean absolute error (MAE), root mean squared error (RMSE), coefficient of determination ($R^2$), and relative absolute error (RAE) given by
\begin{align}
\mbox{RAE} = \frac{\sum_{n=1}^N |\hat{y} - y|}{\sum_{n=1}^N |\bar{y} - y|}, \nonumber
\end{align}
where $\hat{y}$ denotes the inferred speed and $\bar{y}$ the average speed. 

\begin{table}[t]
\centering
\caption{Results for different methods across study area 1 and 2 for a missing ratio of 50\%.}
\label{table:results}
\begin{tabular}{llllll}
\hline
Area     & Method               & MAE   & RMSE  & RAE    & $R^2$    \\
\hline
\multirow{24}{*}{Area 1} & Naive & 4.230 & 6.843 & 72.856 & 0.308 \\
         & ARIMA                & 2.874 & 5.160 & 49.490 & 0.606 \\
         & Lin. Reg. (B)    & 2.492 & 4.369 & 42.911 & 0.718 \\
         & Lin. Reg. (A)    & 2.486 & 4.361 & 42.823 & 0.719 \\
         & Lin. Reg. (B+A) & 2.442 & 4.294 & 42.053 & 0.727 \\
         & kNN (B)		& 2.836 & 4.631 & 48.835 & 0.683 \\
         & kNN (A)		& 2.817 & 4.599 & 48.510 & 0.687 \\
         & kNN (B+A)		& 2.722 & 4.446 & 46.874 & 0.708 \\
         & B-EM (B)		& 3.077 & 5.116 & 52.994 & 0.613 \\
         & B-EM (A)		& 3.058 & 5.069 & 52.671 & 0.620 \\
         & B-EM (B+A)		& 2.713 & 4.641 & 46.730 & 0.682 \\
         & PPCA (B)		& 3.132 & 5.133 & 53.939 & 0.611 \\
         & PPCA (A)		& 3.107 & 5.143 & 53.508 & 0.609 \\
         & PPCA (B+A)		& 2.838 & 4.728 & 48.871 & 0.670 \\
         & Bi-LSTM (B)		& 2.616 & 4.502 & 45.054 & 0.700 \\
         & Bi-LSTM (A)		& 2.595 & 4.513 & 44.693 & 0.699 \\
         & Bi-LSTM (B+A)		& 2.583 & 4.495 & 44.479 & 0.701 \\
         & Indep. GP            & 2.365 & 4.202 & 40.740 & 0.739 \\
         & VARMA (B)        & 2.395 & 4.248 & 41.242 & 0.733 \\
         & VARMA (A)        & 2.395 & 4.248 & 41.244 & 0.733 \\
         & VARMA (B+A)     & 2.377 & 4.217 & 40.946 & 0.737 \\
         & Multi-GP (B)     & 1.900 & 3.599 & 32.719 & 0.809 \\
         & Multi-GP (A)     & \textbf{1.864} & \textbf{3.560} & \textbf{32.102} & \textbf{0.813} \\
         & Multi-GP (B+A)  & 1.951 & 3.786 & 33.604 & 0.788 \\
\hline
\multirow{24}{*}{Area 2} & Naive  & 4.460 & 7.513 & 66.120 & 0.389 \\
         & ARIMA                & 3.770 & 6.647 & 55.890 & 0.522 \\
         & Lin. Reg. (B)    & 2.631 & 4.846 & 39.013 & 0.746 \\
         & Lin. Reg. (A)    & 2.619 & 4.828 & 38.826 & 0.748 \\
         & Lin. Reg. (B+A) & 2.582 & 4.761 & 38.275 & 0.755 \\
         & kNN (B)	& 3.060 & 5.216 & 45.361 & 0.706   \\
         & kNN (A)	& 3.022 & 5.117 & 44.800 & 0.717   \\
         & kNN (B+A)	& 2.981 & 5.067 & 44.200 & 0.722   \\
         & B-EM (B)	& 3.552 & 5.990 & 52.654 & 0.612   \\
         & B-EM (A)	& 3.509 & 5.872 & 52.019 & 0.627   \\
         & B-EM (B+A)	& 3.228 & 5.546 & 47.861 & 0.667   \\
         & PPCA (B)	& 3.752 & 6.314 & 55.619 & 0.569   \\
         & PPCA (A)	& 3.636 & 6.074 & 53.903 & 0.601   \\
         & PPCA (B+A)	& 3.421 & 5.640 & 50.712 & 0.656   \\
         & Bi-LSTM (B)	& 2.729 & 4.934 & 40.459 & 0.737   \\
         & Bi-LSTM (A)	& 2.706 & 4.895 & 40.122 & 0.741 \\
         & Bi-LSTM (B+A)	& 2.704 & 4.852 & 40.090 & 0.745 \\
         & Indep. GP            & 2.482 & 4.611 & 36.795 & 0.770 \\
         & VARMA (B)        & 2.538 & 4.693 & 37.688 & 0.761 \\
         & VARMA (A)        & 2.530 & 4.641 & 37.510 & 0.767 \\
         & VARMA (B+A)     & 2.511 & 4.609 & 37.221 & 0.770 \\
         & Multi-GP (B)     & 2.167 & 4.177 & 32.132 & 0.811 \\
         & Multi-GP (A)     & \textbf{2.104} & \textbf{4.063} & \textbf{31.199} & \textbf{0.821} \\
         & Multi-GP (B+A)  & 2.105 & 4.071 & 32.201 & 0.820 \\
\hline
\end{tabular}
\end{table}

\begin{table}[t]
\centering
\caption{Results for different methods across study area 3 and 4 for a missing ratio of 50\%.}
\label{table:results2}
\begin{tabular}{llllll}
\hline
Area     & Method               & MAE   & RMSE  & RAE    & $R^2$    \\
\hline
\multirow{24}{*}{Area 3} & Naive  & 3.839 & 6.139 & 69.956 & 0.313 \\
         & ARIMA                & 2.147 & 3.741 & 39.124 & 0.745 \\
         & Lin. Reg. (B)    & 2.223 & 3.890 & 40.504 & 0.724 \\
         & Lin. Reg. (A)    & 2.204 & 3.864 & 40.153 & 0.728 \\
         & Lin. Reg. (B+A) & 2.170 & 3.806 & 39.548 & 0.736 \\
         & kNN (B)	& 2.489 & 4.039 & 45.360 & 0.703  \\
         & kNN (A)	& 2.483 & 4.018 & 45.239 & 0.706  \\
         & kNN (B+A)	& 2.400 & 3.906 & 43.735 & 0.722  \\
         & B-EM (B)	& 2.601 & 4.289 & 47.403 & 0.665  \\
         & B-EM (A)	& 2.568 & 4.207 & 46.800 & 0.677  \\
         & B-EM (B+A)	& 2.259 & 3.802 & 41.160 & 0.736\\
         & PPCA (B)	& 2.689 & 4.349 & 48.994 & 0.655  \\
         & PPCA (A)	& 2.675 & 4.301 & 48.748 & 0.663  \\
         & PPCA (B+A)	& 2.428 & 3.971 & 44.244 & 0.712  \\
         & Bi-LSTM (B)	& 2.348 & 4.046 & 42.778 & 0.702  \\
         & Bi-LSTM (A)	& 2.325 & 3.998 & 42.372 & 0.709  \\
         & Bi-LSTM (B+A)	& 2.289 & 3.965 & 41.702 & 0.713  \\
         & Indep. GP            & 2.075 & 3.717 & 37.817 & 0.748 \\
         & VARMA (B)        & 2.085 & 3.698 & 37.998 & 0.751 \\
         & VARMA (A)        & 2.076 & 3.686 & 37.827 & 0.752 \\
         & VARMA (B+A)     & 2.061 & 3.660 & 37.553 & 0.756 \\
         & Multi-GP (B)     & 1.594 & 2.972 & 29.037 & 0.839 \\
         & Multi-GP (A)     & 1.585 & \textbf{2.962} & 28.886 & \textbf{0.840} \\
         & Multi-GP (B+A)  & \textbf{1.579} & 2.969 & \textbf{28.781} & 0.839 \\
\hline
\multirow{36}{*}{Area 4} & Naive & 3.226 & 5.314 & 58.934 & 0.499 \\
         & ARIMA                & 1.971 & 3.466 & 36.008 & 0.787 \\
         & Lin. Reg. (BR)    & 1.909 & 3.533 & 34.876 & 0.778 \\
         & Lin. Reg. (BL)    & 1.835 & 3.426 & 33.528 & 0.792 \\
         & Lin. Reg. (A)     & 1.810 & 3.370 & 33.067 & 0.798 \\
         & Lin. Reg. (BL+A)  & 1.759 & 3.296 & 32.136 & 0.807 \\
         & kNN (BR)		& 2.531 & 4.146 & 46.233 & 0.695\\
         & kNN (BL)		& 2.124 & 3.633 & 38.805 & 0.766\\
         & kNN (A)		& 2.093 & 3.587 & 38.234 & 0.772\\
         & kNN (BL+A)		& 2.031 & 3.484 & 37.101 & 0.784\\
         & EM (BR)		& 2.254 & 3.878 & 39.124 & 0.721\\
         & EM (BL)		& 2.123 & 3.775 & 38.776 & 0.747\\
         & EM (A)		& 2.037 & 3.590 & 37.206 & 0.771\\
         & EM (BL+A)		& 1.740 & 3.221 & 31.790 & 0.816\\
         & PPCA (BR)		& 2.212 & 3.876 & 40.563 & 0.729\\
         & PPCA (BL)		& 2.217 & 3.882 & 40.497 & 0.732\\
         & PPCA (A)		& 2.103 & 3.617 & 38.417 & 0.768\\
         & PPCA (BL+A)		& 1.933 & 3.375 & 35.318 & 0.798\\
         & Bi-LSTM (BR)		& 1.966 & 3.578 & 35.988 & 0.771\\
         & Bi-LSTM (BL)		& 1.951 & 3.534 & 35.641 & 0.778\\
         & Bi-LSTM (A)		& 1.914 & 3.472 & 34.961 & 0.786\\
         & Bi-LSTM (BL+A)		& 1.854 & 3.376 & 33.864 & 0.798\\
         & Indep. GP         & 1.749 & 3.275 & 31.947 & 0.810 \\
         & VARMA (BL)        & 1.780 & 3.310 & 32.525 & 0.805 \\
         & VARMA (BR)        & 1.790 & 3.329 & 32.708 & 0.803 \\
         & VARMA (A)         & 1.776 & 3.305 & 32.450 & 0.806 \\
         & VARMA (BL+A)      & 1.770 & 3.296 & 32.332 & 0.807 \\
         & Multi-GP (BR)     & 1.660 & 3.166 & 30.331 & 0.822 \\
         & Multi-GP (BL)     & 1.256 & 2.630 & 22.937 & 0.877 \\
         & Multi-GP (A)      & 1.191 & \textbf{2.477} & 21.758 & \textbf{0.891} \\
         & Multi-GP (BL+A)  & \textbf{1.190} & 2.486 & \textbf{21.746} & 0.890 \\
\hline
\end{tabular}
\end{table}

Tables~\ref{table:results} and \ref{table:results2} show the obtained results for a missing ratio of 50\%, which corresponds to a reasonable representation of a wide range of traffic speed datasets that are based on probe vehicle data. Other ratios of missing data will be considered in Section~\ref{subsec:missing_ratios}. The notation ``B", ``A" and ``B+A" is used to denote whether information from the road segments before (``B"), after (``A") or both (``B+A") is considered by the model for traffic speeds at the target road segment. Since the study area 4 contains an intersection, we consider two preceding road segments: ``BL" and ``BR". 

As the results show, the proposed multi-output GP approach for jointly modeling related time-series of traffic speeds significantly outperforms all the other methods, leading to improvements in MAE between 15\% and 32\%. Moreover, we can observe that, as expected, the multi-output approaches outperform their independent counterparts (e.g. ``Multi-GP" outperforms ``Indep. GP", and ``VARMA" outperforms ``ARIMA" and ``Lin. Reg."). Similarly, we can verify that GP-based approaches lead to better imputation results when compared to their linear competitors. 

Interestingly, we can observe that, for the proposed multi-output GP approach, jointly modeling the traffic speeds at the target road segment whose missing values we wish to impute, together with the road segment after (``A") produces the best results. Indeed, the results from Table~\ref{table:results} suggest that, for a missing ratio of 50\%, considering information from more than 2 nearby road segments does not lead to improvements for the proposed multi-output GP. We further experimented with different numbers of latent functions $Q$, but this finding was still verified. In fact, the results obtained for the VARMA model, also suggest that further adding information from adjacent road segments only leads to marginal improvements in imputation quality. 

However, the same does not apply to higher ratios of missing observations. Table~\ref{table:results75} shows the results for a missing ratio of 75\%, where, due to clarity and interpretability reasons, we report only the results for the best performing baseline model - VARMA - and the best performing versions of some reference methods from the state of the art. These results clearly show the superior performance of the proposed multi-output GP approach over the best baseline model. However, in this case, we can verify that further adding the preceding road segment to the multi-output GP does indeed lead to better imputation accuracy at the target road segment. This is reasonable, since for very high missing ratios, it becomes more likely that the observations at two consecutive road segments are simultaneously missing, therefore becoming increasingly useful to leverage observations from other related road segments. 

\begin{table}[t]
\centering
\caption{Results for multi-output GP across study areas for a missing ratio of 75\%.}
\label{table:results75}
\begin{tabular}{llllll}
\hline
Area     & Method               & MAE   & RMSE  & RAE    & $R^2$    \\
\hline
\multirow{7}{*}{Area 1} 
& Best kNN & 3.786 & 5.931 & 65.102 & 0.480 \\
& Best PPCA & 3.935 & 6.493 & 67.663 & 0.376 \\
& Best Bi-LSTM & 3.895 & 6.282 & 66.967 & 0.416 \\
& Best VARMA & 3.611 & 5.739 & 62.084 & 0.513 \\
& Multi-GP (B)    & 2.996 & 4.914 & 51.513 & 0.643 \\
& Multi-GP (A)    & 2.951 & 4.884 & 50.729 & 0.647 \\
& Multi-GP (B+A) & \textbf{2.788} & \textbf{4.729} & \textbf{47.942} & \textbf{0.669} \\
\hline
\multirow{7}{*}{Area 2} 
& Best kNN & 4.108 & 6.660 & 61.049 & 0.517 \\
& Best PPCA & 4.421 & 7.164 & 65.700 & 0.441 \\
& Best Bi-LSTM & 4.123 & 6.874 & 61.278 & 0.485 \\
& Best VARMA & 3.829 & 6.282 & 56.905 & 0.570  \\
& Multi-GP (B)    & 3.312 & 5.527 & 49.225 & 0.667 \\
& Multi-GP (A)    & 3.257 & 5.454 & 48.408 & 0.676 \\
& Multi-GP (B+A) & \textbf{3.221} & \textbf{5.385} & \textbf{47.872} & \textbf{0.684} \\
\hline
\multirow{7}{*}{Area 3} 
& Best kNN & 3.357 & 5.198 & 60.722 & 0.515 \\
& Best PPCA & 3.427 & 5.563 & 61.984 & 0.445 \\
& Best Bi-LSTM & 3.446 & 5.467 & 62.317 & 0.464 \\
& Best VARMA & 3.155 & 4.986 & 57.067 & 0.554 \\
& Multi-GP (B)    & 2.544 & 4.177 & 46.004 & 0.687 \\
& Multi-GP (A)    & 2.544 & 4.174 & 46.019 & 0.687 \\
& Multi-GP (B+A) & \textbf{2.326} & \textbf{3.937} & \textbf{42.076} & \textbf{0.722} \\
\hline
\multirow{8}{*}{Area 4} 
& Best kNN & 2.848 & 4.649 & 51.939 & 0.617 \\
& Best PPCA & 2.643 & 4.640 & 48.196 & 0.619 \\
& Best Bi-LSTM  & 2.910 & 4.878 & 53.077 & 0.579 \\
& Best VARMA & 2.843 & 4.687 & 51.852 & 0.611 \\
& Multi-GP (BR)	& 2.586	& 4.305	& 47.165 & 0.672 \\
& Multi-GP (BL)    & 2.089 & 3.721 & 38.102 & 0.755 \\
& Multi-GP (A)    & 2.015 & 3.580  & 36.757 & 0.773 \\
& Multi-GP (BL+A) & \textbf{1.759} & \textbf{3.272} & \textbf{32.075} & \textbf{0.811} \\
\hline
\end{tabular}
\end{table}

\subsection{Effect of the missing ratio}
\label{subsec:missing_ratios}

\begin{figure*}[!t]
	\centering
	\subfloat[Area 1]{\includegraphics[width=0.49\linewidth, trim={.0cm 0 1.9cm 1.3cm},clip]{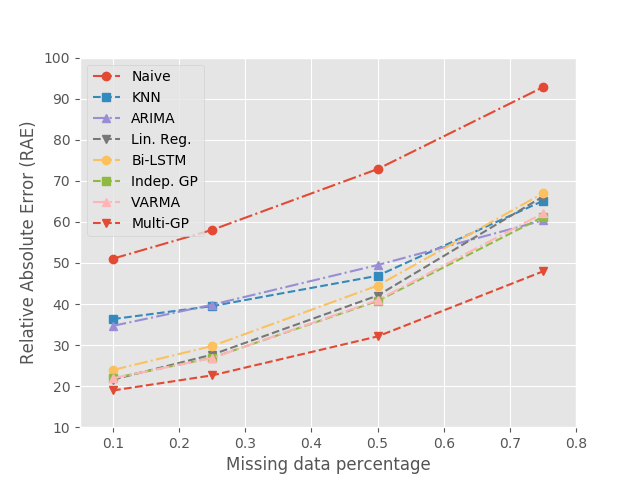}}\hspace{0.25cm}
	\subfloat[Area 2]{\includegraphics[width=0.49\linewidth, trim={.0cm 0 1.9cm 1.3cm},clip]{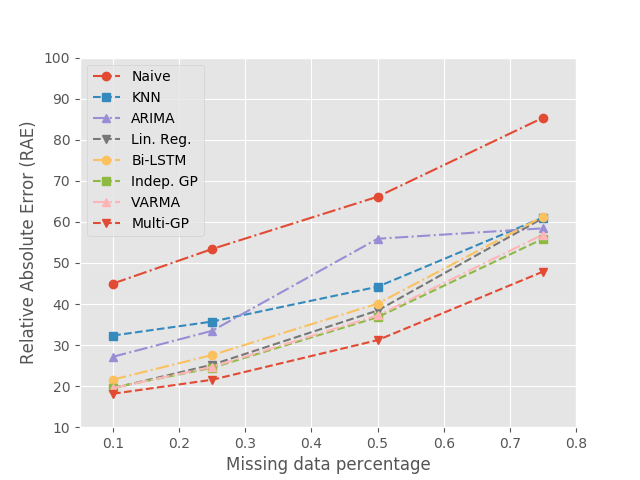}}
	
	\subfloat[Area 3]{\includegraphics[width=0.49\linewidth, trim={.0cm 0 1.9cm 1.3cm},clip]{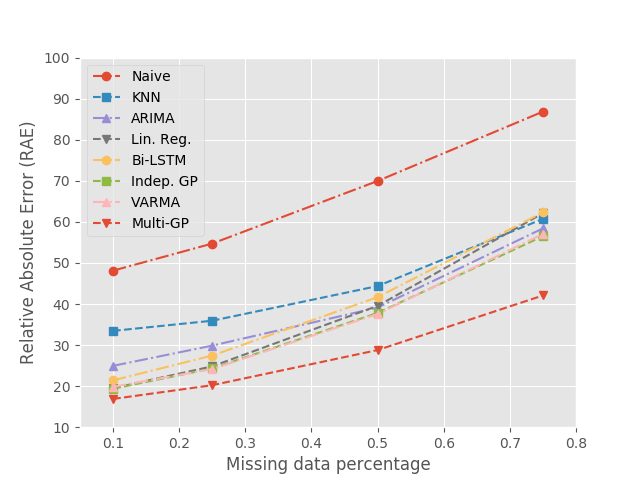}}\hspace{0.25cm}
	\subfloat[Area 4]{\includegraphics[width=0.49\linewidth, trim={.0cm 0 1.9cm 1.3cm},clip]{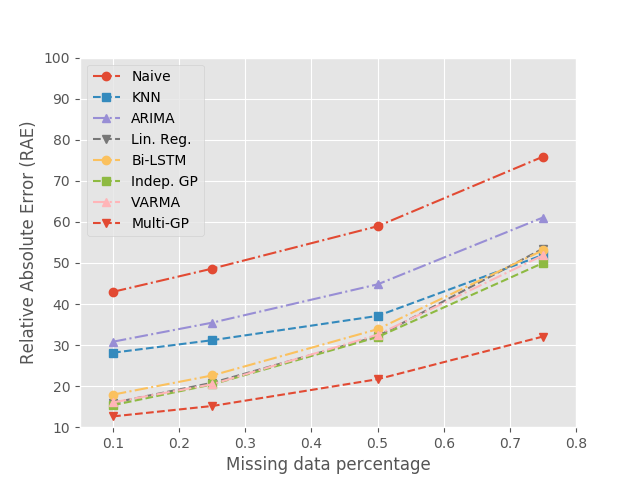}}
	\caption{Results for different missing ratios across different areas.}
	\label{fig:results_plots}
\end{figure*}

In this section, we aim at analyzing the performance of the different imputation approaches for different ratios of missing observations. For this purpose, we performed experiments with the following missing ratios: 10\%, 25\%, 50\% and 75\%. Figure~\ref{fig:results_plots} shows the obtained results for the 4 study areas. In order to keep the plots uncluttered, we report only the results for the best version of each model (``B", ``A" or ``B+A"). 

As expected, the results from Figure~\ref{fig:results_plots} show that, as the missing ratio increases, the accuracy of all imputation methods decreases. However, we can clearly verify that the proposed multi-output GP approach significantly outperforms all the other imputation methods throughout all study areas regardless of the missing ratio. In fact, we can observe that the gap between Multi-GP and the other methods becomes larger for higher missing ratios. This demonstrates the robustness and generalization capabilities of the proposed approach, since for situations when there are very little speed observations available, the multi-output GP model can leverage its knowledge of the temporal structure of the time-series, as well as the complex dependencies between the speeds at nearby road segments, to more accurately extrapolate to periods and locations without speed observations from the mobile devices.

\subsection{Effect of the missing pattern}
\label{subsec:missing_pattern}

Lastly, we evaluated the performance of different methods when subject to different missing patterns. In order to perform such evaluation, we followed an approach similar to one recently proposed by Tak et. al. \cite{tak2016data}, where missing data occurs in blocks. To make this setting even more realistic and generate ``bursts" of missing observations of varying length, we implemented a finite-state machine (FSM) with two states: observed and missing. At each time step $t$, the FSM then has some probability $p(m|o)$ of transitioning from the observed state $o$ to the missing state $m$, and it can stay at that missing state with some probability $p(m|m) = 1 - p(o|m)$. Using this procedure, we simulated missing observations for all the study areas. We considered two scenarios: one where $p(m|o) = 0.25$ and $p(m|m) = 0.75$, which corresponds to a overall missing rate of approximately 49\%, and another scenario where $p(m|o) = 0.5$ and $p(m|m) = 0.8$, which corresponds to a overall missing rate of approximately 69\%. Tables~\ref{table:results2575} and \ref{table:results5080} show, respectively, the results for both these scenarios. For the sake of clarity and interpretability, we report only the results for the best performing baseline model - VARMA - and the best performing versions of some reference methods from the state of the art. Contrarily to the experiments from the previous sections, this experiment setup corresponds to scenarios where the data is not missing  completely at random (MCAR), but rather to a scenario where there is a temporal dependency between likelihood of an observation being missing. 

From the obtained results in Tables~\ref{table:results2575} and \ref{table:results5080}, we can observe that the general trend from the earlier experiments can still be verified, with the proposed multi-output GP approach significantly outperforming all the other baselines methods from the state of the art. Indeed, in some cases the improvements of the proposed Multi-GP and the other baselines methods is even more substantial. Since the missing data now occurs for long continuous periods (``bursts"), this experiments further show the ability of multi-output GPs in modeling the correlations between nearby road segments, and in capturing the spatio-temporal patterns in the observed data in order to extrapolate over long temporal ranges and produce significantly more accurate imputations. Furthermore, the robustness of Multi-GP to high missing rates becomes once again evident. 

\begin{table}[t]
\centering
\caption{Results across study areas for missing patterns simulated with a FSM with probabilities $p(m|o) = 0.25$ and $p(m|m) = 0.75$.}
\label{table:results2575}
\begin{tabular}{llllll}
\hline
Area     & Method               & MAE   & RMSE  & RAE    & $R^2$    \\
\hline
\multirow{7}{*}{Area 1} 
& Best kNN & 3.468   & 5.410   & 60.817  & 0.540 \\
& Best PPCA & 3.288   & 5.387   & 57.666  & 0.544 \\
& Best Bi-LSTM & 3.735   & 6.087   & 65.506  & 0.418 \\
& Best VARMA & 3.354   & 5.492   & 58.823  & 0.526 \\
& Multi-GP (B)    & 2.493   & 4.385  & 43.719  & 0.698 \\
& Multi-GP (A)    & 2.459   & 4.351  & 43.127  & 0.703 \\
& Multi-GP (B+A) & \textbf{2.259} & \textbf{4.131} & \textbf{39.614} & \textbf{0.732} \\ 
\hline
\multirow{7}{*}{Area 2} 
& Best kNN & 3.836   & 6.217   & 57.758  & 0.566 \\
& Best PPCA & 3.991   & 6.506   & 60.096  & 0.524 \\
& Best Bi-LSTM & 3.853   & 6.432   & 58.022  & 0.535 \\
& Best VARMA & 3.530   & 5.977   & 53.150  & 0.599  \\
& Multi-GP (B)    & 2.930   & 5.073   & 44.125  & 0.711 \\
& Multi-GP (A)    & \textbf{2.833} & \textbf{5.026} & \textbf{42.658} & \textbf{0.716} \\
& Multi-GP (B+A) &  2.992   & 5.179   & 45.059  & 0.699 \\
\hline
\multirow{7}{*}{Area 3} 
& Best kNN & 3.100   & 4.783   & 56.860  & 0.577 \\
& Best PPCA & 2.849   & 4.584   & 52.246  & 0.611 \\
& Best Bi-LSTM & 3.316   & 5.270   & 60.808  & 0.486 \\
& Best VARMA & 2.922   & 4.655   & 53.588  & 0.599 \\
& Multi-GP (B)    & 2.155   & 3.694   & 39.523  & 0.748 \\
& Multi-GP (A)    & 2.141   & 3.650   & 39.267  & 0.753 \\
& Multi-GP (B+A) & \textbf{2.023} & \textbf{3.513} & \textbf{37.110} & \textbf{0.772} \\ 
\hline
\multirow{8}{*}{Area 4} 
& Best kNN & 2.663   & 4.389   & 48.802  & 0.656 \\
& Best PPCA & 2.275   & 3.963   & 41.693  & 0.720 \\
& Best Bi-LSTM  & 2.780   & 4.657   & 50.950  & 0.613 \\
& Best VARMA & 2.661   & 4.537   & 48.767  & 0.633 \\
& Multi-GP (BR)	& 1.940   & 3.680   & 35.566  & 0.758 \\
& Multi-GP (BL)    & 1.717   & 3.352   & 31.468  & 0.800 \\
& Multi-GP (A)    & 1.657   & 3.237   & 30.366  & 0.813 \\
& Multi-GP (BL+A) & \textbf{1.479} & \textbf{3.005} & \textbf{27.118} & \textbf{0.839} \\ 
\hline
\end{tabular}
\end{table}

\begin{table}[t]
\centering
\caption{Results across study areas for missing patterns simulated with a FSM with probabilities $p(m|o) = 0.5$ and $p(m|m) = 0.8$.}
\label{table:results5080}
\begin{tabular}{llllll}
\hline
Area     & Method               & MAE   & RMSE  & RAE    & $R^2$    \\
\hline
\multirow{7}{*}{Area 1} 
& Best kNN & 4.003   & 6.177   & 69.658  & 0.414 \\
& Best PPCA & 4.000   & 6.501   & 69.607  & 0.351 \\
& Best Bi-LSTM & 4.092   & 6.487   & 71.193  & 0.354 \\
& Best VARMA & 3.762   & 5.926   & 65.467  & 0.461 \\
& Multi-GP (B)    & 3.037   & 4.986   & 52.843  & 0.619 \\
& Multi-GP (A)    & 2.992   & 4.929   & 52.059  & 0.627 \\
& Multi-GP (B+A) & \textbf{2.863} & \textbf{4.876} & \textbf{49.821} & \textbf{0.635} \\
\hline
\multirow{7}{*}{Area 2} 
& Best kNN & 4.344   & 6.932   & 64.958  & 0.468 \\
& Best PPCA & 4.534   & 7.415   & 67.794  & 0.391 \\
& Best Bi-LSTM & 4.286   & 7.015   & 64.087  & 0.455 \\
& Best VARMA & 3.941   & 6.397   & 58.923  & 0.547  \\
& Multi-GP (B)    & 3.401   & 5.626   & 50.845  & 0.649 \\
& Multi-GP (A)    & 3.354   & 5.609   & 50.150  & 0.652 \\
& Multi-GP (B+A) & \textbf{3.350} & \textbf{5.542} & \textbf{50.083} & \textbf{0.660} \\   
\hline
\multirow{7}{*}{Area 3} 
& Best kNN & 3.578   & 5.460   & 64.996  & 0.459 \\
& Best PPCA & 3.684   & 5.758   & 66.913  & 0.399 \\
& Best Bi-LSTM & 3.652   & 5.719   & 66.325  & 0.407 \\
& Best VARMA & 3.281   & 5.113   & 59.589  & 0.526 \\
& Multi-GP (B)    & 2.642   & 4.314   & 47.981  & 0.663 \\
& Multi-GP (A)    & 2.622   & 4.246   & 47.622  & 0.673 \\
& Multi-GP (B+A) & \textbf{2.414} & \textbf{4.044} & \textbf{43.842} & \textbf{0.704} \\  
\hline
\multirow{8}{*}{Area 4} 
& Best kNN & 3.093   & 4.982   & 56.714  & 0.555 \\
& Best PPCA & 2.803   & 5.174   & 51.400  & 0.520 \\
& Best Bi-LSTM  & 3.164   & 5.218   & 58.021  & 0.512 \\
& Best VARMA & 2.962   & 4.854   & 54.303  & 0.578 \\
& Multi-GP (BR)	& 2.556   & 4.376   & 46.863  & 0.657 \\
& Multi-GP (BL)    & 2.179   & 3.884   & 39.960  & 0.730 \\
& Multi-GP (A)    & 2.110   & 3.745   & 38.690  & 0.749 \\
& Multi-GP (BL+A) & \textbf{1.816} & \textbf{3.395} & \textbf{33.296} & \textbf{0.793} \\ 
\hline
\end{tabular}
\end{table}

\section{Conclusion}
\label{sec:conclusion}

This article proposed the use of multi-output Gaussian processes for crowdsourced traffic speed data imputation. Thanks to their Bayesian formalism, multi-output GPs are able to handle the observation noise that is inherent to crowdsourced traffic data, while its multi-output generalization based on convolution processes allows them to capture and exploit complex spatial dependencies and correlations between nearby road segments to improve imputation accuracy. We empirically evaluated the proposed approach using 6 months of crowdsourced traffic data from Copenhagen. As the obtained results show, the proposed approach significantly outperforms an extensive list of state-of-the-art imputation methods. Moreover, we were able to verify and quantify the improvements obtained by considering the spatial correlations with nearby road segments, and identify the most influential ones. Finally, our results demonstrate that just by considering the complex spatial dependency with the subsequent road segment, the proposed approach is able to obtain very substantial improvements over all the state-of-the-art methods considered. 



%



\section*{Acknowledgment}

The research leading to these results has received funding from the People Programme (Marie Curie Actions) of the European Union’s Horizon 2020 research and innovation programme under the Marie Sklodowska-Curie Individual Fellowship H2020-MSCA-IF- 2016, ID number 745673, and from the People Programme (Marie Curie Actions) of the European Union’s Seventh Framework Programme (FP7/2007-2013) under REA grant agreement no. 609405 (COFUNDPostdocDTU). The authors would also like to thank Google for proving access to the data used in this work.

\ifCLASSOPTIONcaptionsoff
  \newpage
\fi



\bibliographystyle{IEEEtran}
\bibliography{multi-gp-imputation}
%



%


\begin{IEEEbiography}[{\includegraphics[width=1in,height=1.25in,clip,keepaspectratio]{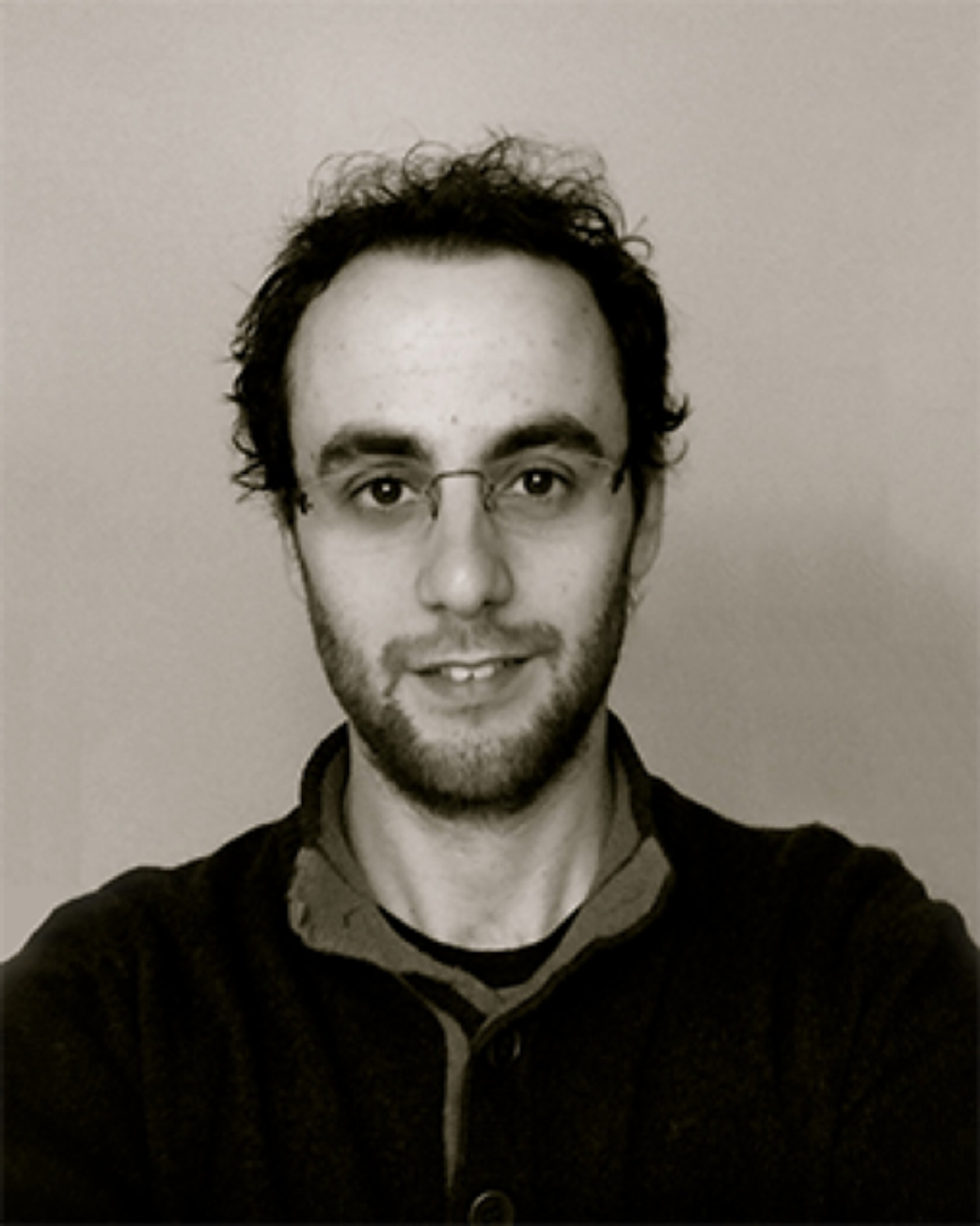}}]{Filipe~Rodrigues}
is Postdoctoral Fellow at Technical University of Denmark, where he is working on machine learning models for understanding urban mobility and the behaviour of crowds, with emphasis on the effect of special events in mobility and transportation systems. He received a Ph.D. degree in Information Science and Technology from University of Coimbra, Portugal, where he developed probabilistic models for learning from crowdsourced and noisy data. His research interests include machine learning, probabilistic graphical models, natural language processing, intelligent transportation systems and urban mobility. 
\end{IEEEbiography}

\begin{IEEEbiography}[{\includegraphics[width=1in,height=1.25in,clip,keepaspectratio]{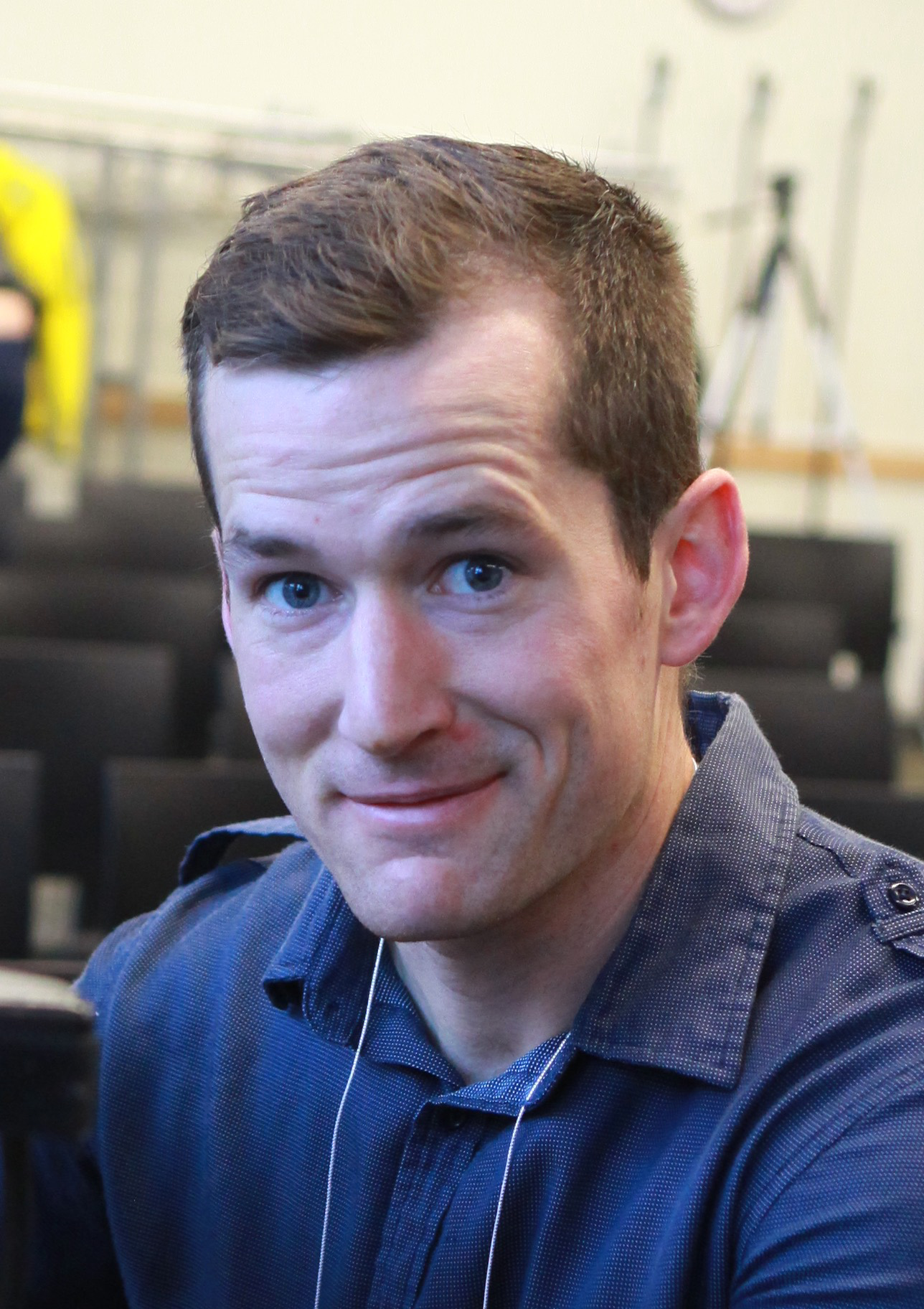}}]{Kristian-Henrickson}
received a B.S. in Civil Engineering from the University of Idaho in Moscow, ID in 2013, and M.S. in Civil and Environmental Engineering from the University of Washington in Seattle, WA in 2014. He is currently pursuing the Ph.D. degree in Civil and Environmental Engineering at the University of Washington. From 20012 to 2013, he was an undergraduate Research Assistant with the National Institute for Advanced Traffic Safety at the University of Idaho. Since 2013, he has worked as a Research and Teaching Assistant at the University of Washington STAR Laboratory. His research interest includes transportation data quality issues, machine learning, and transportation safety.
\end{IEEEbiography}

\begin{IEEEbiography}[{\includegraphics[width=1in,height=1.25in,clip,keepaspectratio]{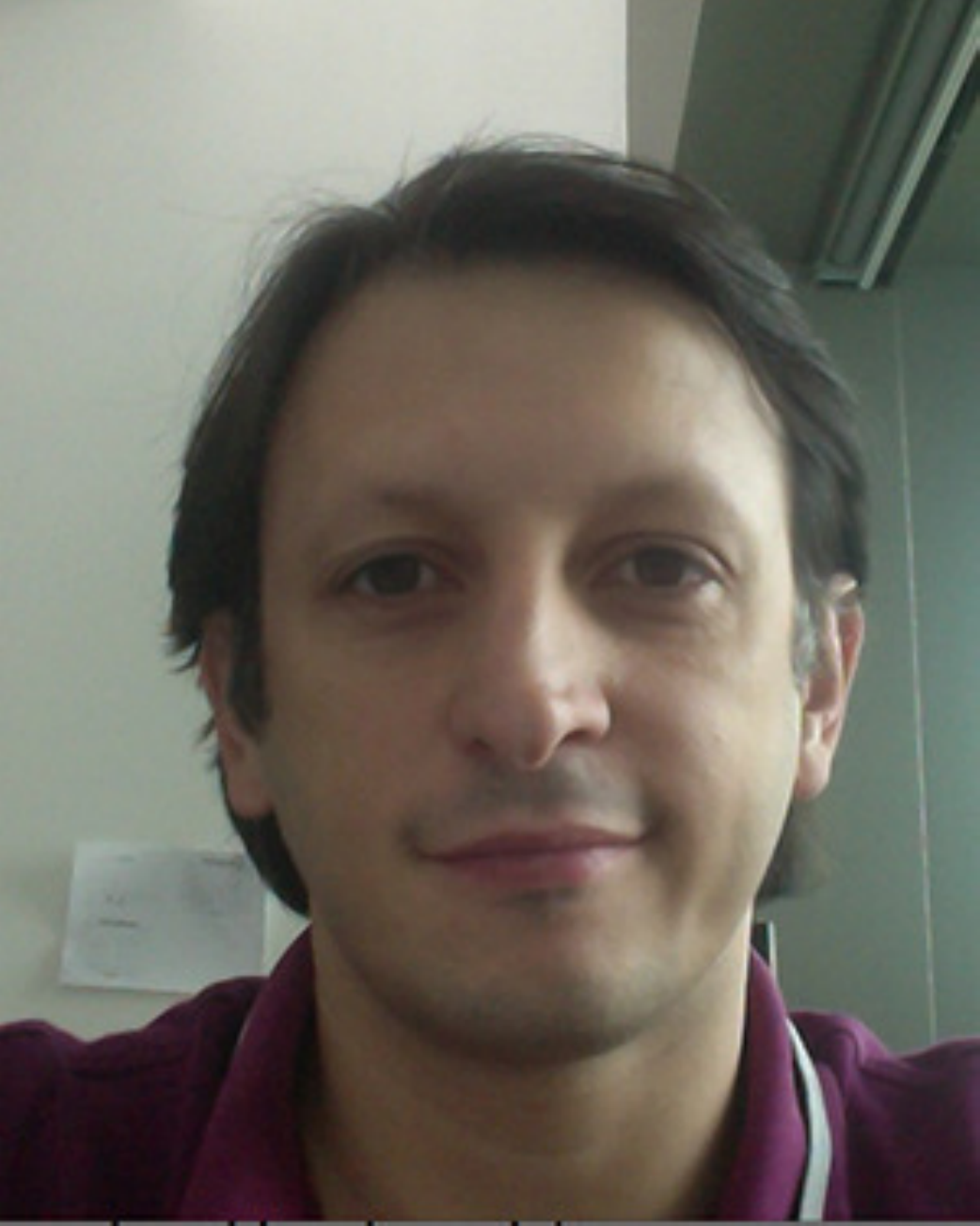}}]{Francisco~C.~Pereira}
is Full Professor at the Technical University of Denmark (DTU), where he leads the Smart Mobility research group. His main research focus is on applying machine learning and pattern recognition to the context of transportation systems with the purpose of understanding and predicting mobility behavior, and modeling and optimizing the transportation system as a whole. He has Master's (2000) and Ph.D. (2005) degrees in Computer Science from University of Coimbra, and has authored/co-authored over 70 journal and conference papers in areas such as pattern recognition, transportation, knowledge based systems and cognitive science. Francisco was previously Research Scientist at MIT and Assistant Professor in University of Coimbra. He was awarded several prestigious prizes, including an IEEE Achievements award, in 2009, the Singapore GYSS Challenge in 2013, and the Pyke Johnson award from Transportation Research Board, in 2015.
\end{IEEEbiography}




\end{document}